\documentclass[runningheads]{mybi}
\usepackage{graphicx}
\usepackage{subfigure}
\usepackage{amsmath}
\usepackage{amssymb}

\usepackage{booktabs}
\usepackage{threeparttable}
\usepackage{multicol}  
\usepackage{multirow} 
\renewcommand{\arraystretch}{1.5}
\usepackage{color}
\usepackage{xspace}
\newcommand{\method}{M2DAN\xspace}
\usepackage{ragged2e}

\def\ie{\emph{i.e.,~}}
\def\eg{\emph{e.g.,~}}

\def\wrt{\emph{w.r.t.~}}

\def\newl{\vspace{0.05in}}

\usepackage{pifont}
\newcommand{\cmark}{\ding{51}}%

\definecolor{chenyaofo}{RGB}{17,120,100}
\def\cyf{\textcolor{black}}

\definecolor{subyellow}{RGB}{184,134,11}

\definecolor{random}{RGB}{122,11,180}

\definecolor{fan}{RGB}{181,68,52}

\definecolor{qiu}{RGB}{150,100,100}

\begin{document}
%
\title{Multi-Scale Multi-Target Domain Adaptation \\for Angle Closure Classification}
\titlerunning{Multi-Scale Multi-Target Domain Adaptation}
%
\author{Zhen Qiu$^{1,5}$ \and
Yifan Zhang$^{2}$\and
Fei Li$^3$\and
Xiulan Zhang$^3$\and
\\
Yanwu Xu$^{4}$\and
Mingkui Tan$^{1,6}$\thanks{Corresponding author.}}
\authorrunning{Z. Qiu et al.}
%
\institute{$^1$South China University of Technology,
$^2$National University of Singapore,\\
$^3$Sun Yat-sen University,
$^4$Baidu Inc.,
$^5$Pazhou Laboratory, \\$^6$Key Laboratory of Big Data and Intelligent Robot, Ministry of Education\\}

\makeatletter
\renewcommand*{\@fnsymbol}[1]{\ensuremath{\ifcase#1\or *\or \dagger\or \ddagger\or
		\mathsection\or \mathparagraph\or \|\or **\or \dagger\dagger
		\or \ddagger\ddagger \else\@ctrerr\fi}}
\makeatother

\maketitle              
\begin{abstract}
Deep learning (DL) has made significant progress in angle closure classification with anterior segment optical coherence tomography (AS-OCT) images. These AS-OCT images are often acquired by different imaging devices/conditions, which results in a vast change of underlying data distributions (called “data domains”). Moreover, due to practical labeling difficulties, some domains (\eg devices) may not have any data 
labels. As a result,  
deep models trained on one specific domain (\eg a specific device) are difficult to adapt to and thus may perform poorly on other domains (\eg other devices). To address this issue, we present a multi-target domain adaptation  paradigm to transfer a model trained on one labeled source domain to multiple unlabeled target domains. Specifically, we propose a novel Multi-scale Multi-target Domain Adversarial Network (M2DAN) for angle closure classification. M2DAN conducts multi-domain adversarial learning for extracting  domain-invariant features and develops a multi-scale module for capturing local and global information of AS-OCT images. Based on these domain-invariant features at different scales, the deep model trained on the source domain is able to classify angle closure on multiple target domains even without any annotations in these domains. Extensive experiments on a real-world AS-OCT dataset demonstrate  the effectiveness of the proposed method.

\keywords{Angle Closure Classification  \and Unsupervised Multi-target Domain Adaptation \and Anterior Segment Optical Coherence Tomography.}
\end{abstract}
\section{Introduction}
\label{sect1}
\begin{figure}[t] 
\vspace{-0.1in}
\label{figdata}
\centering
\subfigure[AS-I (open angle)]{
\includegraphics[width=5cm]{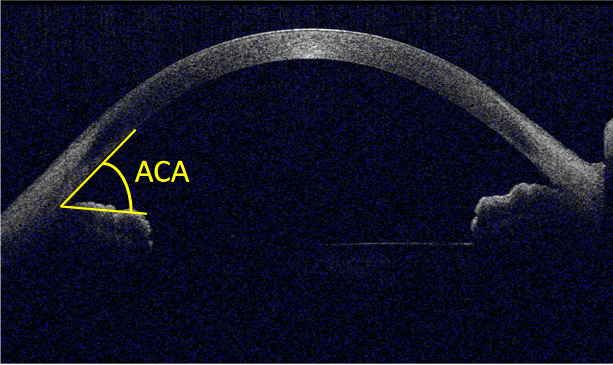}
\label{open}
}
\quad
\subfigure[AS-I (narrow angle)]{
\includegraphics[width=5cm]{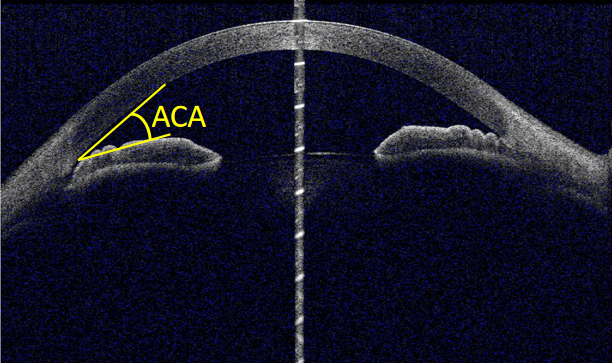}
\label{narrow}
}
\quad
\subfigure[AS-II]{
\includegraphics[width=5cm]{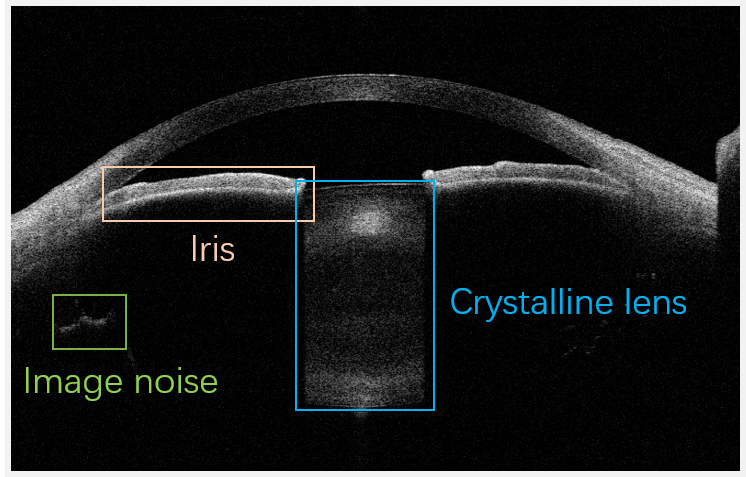}
\label{external}
}
\quad
\subfigure[AS-III]{
\includegraphics[width=5cm]{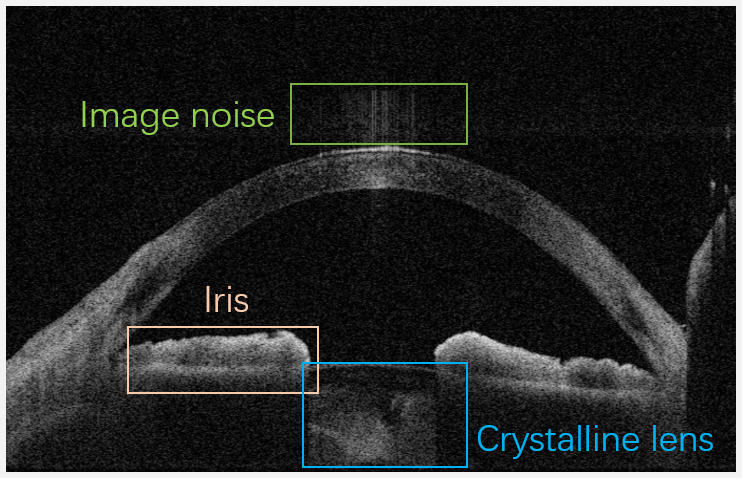}
\label{casia2}
} 
 \vspace{-0.1in}
\caption{Illustration of different types of anterior chamber angle (ACA) and different data domains. Specifically, ACA consists of two categories: open angle (a) and narrow angle (b). In addition, different imaging devices/techniques may result in different data domains of anterior segment (AS) images, \eg  AS-I (a-b), AS-II (c) and AS-III (d), which differ in terms of the crystalline lens~\cite{dubbelman2005change}, image noises~\cite{azzeh2018salt}, and image resolutions.}
\vspace{-0.1in}
\end{figure}

Glaucoma is the foremost cause of irreversible blindness~\cite{quigley2006number,thylefors1995global}. Since the vision loss is irreversible, early detection and precise diagnosis for glaucoma are essential to  vision preservation. A common type of glaucoma is angle closure, where the anterior chamber angle (ACA) is narrow  as shown in Fig.~\ref{narrow}. Such an issue leads to blockage of drainage channels that results in pressure on the optic nerve~\cite{fu2018multi}.
To identify this, anterior segment optical coherence tomography (AS-OCT) has been shown an effective approach for the evaluation of the ACA structure~\cite{leung2011anterior} and is thus widely used for angle closure classification~\cite{hao2021angle,hao2021hybrid}.

Despite the success of deep learning in computer-aided diagnosis, it hinges on massive annotated images for training~\cite{litjens2017survey}.
Thanks to AS-OCT devices and hence a growing number of labeled AS-OCT data for deep model training, remarkable performance has been achieved on angle closure classification~\cite{fu2020age,niwas2016automated}.
However, different imaging devices/conditions intrinsically lead to a vast change  of  underlying  data  distributions~\cite{zhang2019whole}, which means that the AS-OCT images may come from different  ``domains".  As a result,  
deep models trained on one domain (\eg a specific device) can hardly generalize  to other domains~\cite{niu2022efficient} (\eg other devices).
More critically, it is  impractical to customize deep models for each domain, since annotation costs for such specific-customized images are inevitably expensive. Note that due to labeling difficulties, we may not have any labeled data for some  domains.

To solve this issue, one may explore unsupervised domain adaptation~\cite{benaim2017one,ghafoorian2017transfer,liu2020duda,Qiu2021CPGA,tzeng2017adversarial,tzeng2014deep,wang2018lstn,zhou2019adversarial}, which leverages  the  labeled  data  on a source domain  to improve the performance  on an unlabeled target domain. 
Most existing methods~\cite{bousmalis2016domain,zhang2019whole,zhang2019collaborative,ren2018adversarial,Lin2022ProCA} focus on pair-wise adaptation from one source domain to one target domain. 
However, in angle closure classification tasks, AS-OCT images are often acquired via diverse imaging devices (\eg CASIA-I or CASIA-II), imaging conditions (\eg light or dark environment) and preprocessing techniques. In other words, doctors need to classify AS-OCT images from different domains. Therefore, it is more practical to study multi-target domain adaptation~\cite{gholami2020unsupervised,peng2019domain,yang2019domain} for angle closure classification.
To be specific, multi-target domain adaptation aims to leverage  the  annotations on one source domain to improve the performance of multiple unlabeled target domains.
Despite the importance,  it remains largely unexplored for medical images analysis, especially in angle closure classification.

Multi-target domain adaptation with AS-OCT image poses two challenges. The first challenge is the domain discrepancies among multiple domains, which results from different imaging devices, imaging conditions, and preprocessing techniques. As shown in Fig.~\ref{figdata}, the AS-OCT images from different domains may differ in terms of the crystalline lens~\cite{dubbelman2005change}, image noises~\cite{azzeh2018salt} and image resolutions. As a result, directly applying the model trained on the source domain tends to perform poorly on the multi-target domains. 
The second challenge is how to capture both  local  and global information of AS-OCT images for angle closure classification. In practice, ophthalmologists classify angle closure based on both local information (\eg anterior chamber angle (ACA) and iris curvature) and global information (\eg anterior chamber width  and cornea structure)~\cite{fu2017segmentation}.
However, 
most  deep neural networks (DNNs) tend to learn global features without paying attention to fine-grained information of the images, \eg  ACA in AS-OCT images.
Since the measurement of small regions (\eg trabecular iris angle and angle opening distance~\cite{fu2017segmentation}) in ACA is highly important for this task, it is difficult for existing DNN models to effectively classify angle closure.
As a result, most existing DNN-based unsupervised domain adaptation methods may fail to handle such a challenging task.

To handle the two challenges, we explore multi-domain adversarial learning and  multi-scale feature extraction  for angle closure classification. Specifically, to alleviate the domain discrepancies, we resort to domain adversarial learning, which is one of the mainstream paradigms for pair-wise unsupervised domain adaptation~\cite{zhang2019whole,zhang2019collaborative}. Meanwhile, since there exists low contrast and vast noise in local regions of AS-OCT images (\eg trabecular iris angle and angle opening distance~\cite{fu2017segmentation}),   capturing fine-grained information with a fixed scale of neural filter is   intractable. Therefore, we propose to  explore multi-scale convolutional filters for promoting fine-grained representation extraction~\cite{fu2018multi,zhang2017alignedreid}.
Following these ideas, we present a novel Multi-scale Multi-target Domain Adversarial Network (\method). In \method,  a  new  multi-scale module is developed to capture global and local information of AS-OCT images. Such a module consists of three convolutional branches with different filter sizes, which are used to extract  multi-scale features. Meanwhile, \method conducts multi-domain adversarial learning for each convolutional branch separately,  so that it can  learn domain-invariant features at different scales. Based on the extracted multi-scale domain-invariant features, the classifier trained on the source domain is able to conduct angle closure classification effectively on the multiple target domains. 

We summarize the main contributions of this paper as follows:
\begin{itemize}
\item We propose a novel Multi-scale Multi-target Domain Adversarial  Network for angle closure classification. By exploring a new multi-scale scheme and multi-domain adversarial learning, the proposed method is able to learn multi-scale domain-invariant features and effectively classify the angle closure.   
\item To the best of our knowledge, our work is the first to study multi-target domain adaptation for angle closure classification, which enhances the applications of deep learning in early detection and precise diagnosis for angle closure glaucoma.
\item Extensive experiments demonstrate the effectiveness and superiority of the proposed method on a real-world anterior segment optical coherence tomography dataset with three domains. 
\end{itemize}

\section{Method}
\subsection{Problem Definition}
We consider two practical challenges in AS-OCT based angle closure classification task: 1) the distribution changes of different domains (\eg devices); 2) the lack of labeled data for multiple domains. We tackle them by adapting a model learned on a source domain to $B$ target domains. In total we consider ($B$+$1$) domains.
For convenience, we introduce a domain label vector $\mathbf{d}_{i} \in \{0,1\}^{B+1}$ to indicate the domain labels of each sample. Then, let $\mathcal{D}_s\small{=}\{\mathbf{x}_i, \mathbf{y}_{i},\mathbf{d}_{i}\}_{i=1}^{n_s}$ be the source domain data and $\mathcal{D}_t=\{\mathbf{x}_{j}, \mathbf{d}_{j}\}_{j=1}^{n_t}$ be the unlabeled data collected from $B$ target domains, where $\mathbf{y}_{i}$ denotes the class label of source domain data, and $n_s$ and $n_t$ denote the numbers of samples in $\mathcal{D}_s$ and $\mathcal{D}_t$, respectively.


Note that given a specific task,  all domains share the same label space, but only the source domain data are labeled.  
The primary goal of this paper is  to learn a well-performed deep neural network for multiple target domains, using both labeled source samples and unlabeled target samples.
Unfortunately, since we have no labeled target data, how to conduct effective domain adaptation to multiple target domains is very challenging.


\begin{figure}[t]
\vspace{-0.1in} 
\includegraphics[width=\textwidth]{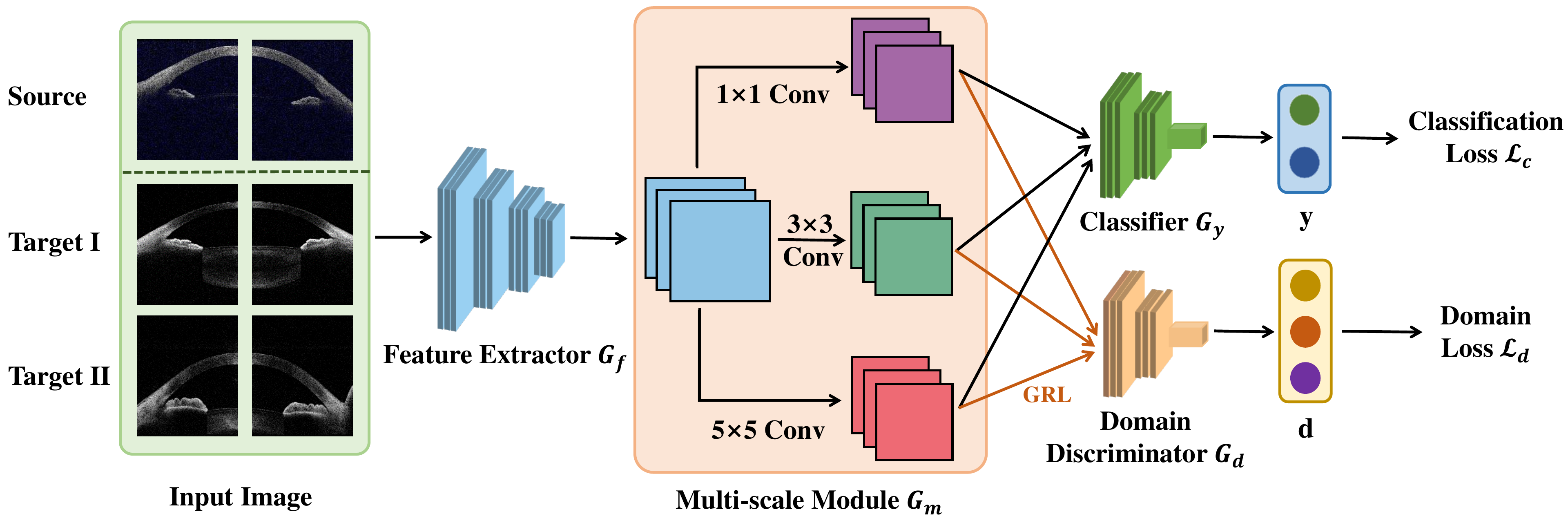}
\caption{The scheme of Multi-scale Multi-target Domain Adversarial Network. Specifically, the multi-scale module consists of three convolutional branches with different filter sizes. Then, we use a shared domain discriminator to distinguish features from each branch separately for scale-aware domain adaptation, while we concatenate all features for classification. We implement domain adversarial training through Gradient Reverse Layer (GRL)~\cite{ganin2014unsupervised}.}
\label{structure}
\vspace{-0.1in}
\end{figure}

\subsection{Multi-scale Multi-target Domain Adversarial Network}

To enforce effective domain adaptation from the source domain $\mathcal{D}_s$ to the multiple target domain $\mathcal{D}_t$, we address the challenges from two aspects: (1) we seek to alleviate the domain discrepancies among multiple domains with  domain adversarial learning; (2) Note that beyond the distribution changes,  the $B+1$ domains also share some intrinsic properties as they are dealing with the same task, \eg angle closure classification. We thus exploit both local and global information of AS-OCT images for angle closure classification with a  multi-scale scheme. 
Given the above motivations,  we propose a multi-scale multi-target domain adversarial network (\method).
As shown in Fig.~\ref{structure}, \method  consists of four components: a feature extractor $G_{f}$ and a multi-scale module $G_{m}$, a classifier $G_{y}$ for task prediction, and a domain discriminator $G_{d}$ to discriminate  multi-scale features of images from different domains. To be specific, the multi-scale module $G_{m}$ is developed to  extract features with multi-scale information.

In the angle closure classification, both local information (\eg iris curvature and angle opening distance ) and global information (\eg  cornea structure~\cite{fu2018multi}) plays important roles. The local information can be used to obtain several major clinical measurements for clinical diagnosis~\cite{nongpiur2013classification}, while the global cornea structure offers various cues associated with risk factors for angle-closure glaucoma~\cite{fu2018multi}. To capture them, the multi-scale module consists of three parallel convolutional branches with different filter sizes (\ie $1 \small{\times} 1$, $3 \small{\times} 3$ and $5 \small{\times} 5$) to extract features from different scales. Here, all feature maps have the same spatial size through padding. We then send the feature maps at different scales into the domain discriminator separately for multiple domain adaptation, while we concatenate all these features for classification.
In this way, we are able to extract domain-invariant features with multi-scale information, which helps the classifier to make more accurate predictions on the target domains. 

To train \method, we employ the following two strategies. 
First, inspired by most domain adaptation methods~\cite{bousmalis2016domain,zhang2019whole,peng2019domain}, we adopt domain adversarial learning to enforce the multi-scale feature extractor $G_{m}\circ G_{f}$ to capture multi-scale domain-invariant features, so that the discrepancy among multiple domains is minimized. To be specific, a shared domain discriminator $G_{d}$ is trained to adequately distinguish features of images from different domains by minimizing a domain loss $\mathcal{L}_{d}$. Meanwhile,  
$G_{m}\circ G_{f}$ is trained to confuse the domain discriminator by maximizing $\mathcal{L}_{d}$.  Note that, domain adversarial learning is applied to each branch of the multi-scale module separately for learning multi-scale domain-invariant features. Second, we train the  backbone network $(G_f,G_m,G_y)$ via a classification loss $\mathcal{L}_c$ to make it imbalance-aware and discriminative. 
The overall training of \method is to solve the following problem:

\begin{align}
 \min_{\theta_{f},\theta_{m},\theta_{y}}\max_{\theta_d}~\underbrace{-\alpha\mathcal{L}_{d}(\theta_{f},\theta_{m},\theta_{d})}_{\text{domain loss}} ~~ +  \underbrace{\lambda\mathcal{L}_{c}(\theta_{f},\theta_{m},\theta_{y})}_{\text{classification loss}} 
\end{align}
where $\theta_{f}, \theta_{m}, \theta_{y}, \theta_{d}$ indicate the parameters of $G_{f}$, $G_{m}$, $G_{y}$ and $G_{d}$, respectively. Moreover,  $\alpha$ and $\lambda$ denote the trade-off parameters for different losses. In next sections, we will detail the domain loss $\mathcal{L}_{d}$ and the classification loss $\mathcal{L}_{c}$.

\subsection{Domain Loss for Multi-target Domain Adaptation}
\label{ds} 
Diverse imaging devices and techniques intrinsically result in discrepancies among image domains. In practice, doctors often need to classify angle closure based on AS-OCT images from multiple target domains. However, existing unsupervised domain adaptation for medical images mainly focuses on two domains and cannot handle this practical problem. 

To solve this, inspired by multi-class classification via cross-entropy, we conduct multi-target domain adaptation via a multi-domain loss  as follows:
\begin{equation}
\mathcal{L}_{d}(\theta_{f},\theta_{m},\theta_{d}) = -\frac{1}{n} \sum_{i=1}^{n} \mathbf{d}_{i}^{\top}\mathrm{log}(\mathbf{\hat{d}}_i),
\end{equation} 
where $\mathbf{\hat{d}}_i\small{=}G_{d}(G_m(G_{f}(\mathbf{x}_{i})))$ denotes the prediction of the domain discriminator \wrt $\mathbf{x}_{i}$, and $n$ denotes the overall number of data. Moreover, since different branches in the multi-scale module share the same domain discriminator, we use the same domain loss for them without explicitly mentioning branches. In this way, \method is able to capture  domain-invariant features at different scales. 

\subsection{Classification Loss for Angle Closure Classification}
\label{cl}
For the task of angle closure classification, we can adopt any classification losses to train the network, \eg cross-entropy. Nevertheless, since the class-imbalanced issue commonly exist in this task, we use the focal loss~\cite{lin2017focal} as follows: 
\begin{equation}
\label{focal-loss}
\mathcal{L}_{fo}(\theta_{f},\theta_{m},\theta_{y}) = -\frac{1}{n_{s}} \sum_{i=1}^{n_{s}} \mathbf{y}_{i}^{\top}\big((1-\mathbf{\hat{y}}_{i})^{\gamma} \odot \mathrm{log}(\mathbf{\hat{y}}_{i})\big),
\end{equation} 
where $\mathbf{\hat{y}}_{i}\small{=}G_{y}(G_{m}(G_{f}(\mathbf{x}_{i}))$ denotes the prediction of the classifier \wrt $\mathbf{x}_{i}$, and $n_{s}$ denotes the number of \textbf{labeled source samples}.
Moreover, $\odot$ denotes the element-wise product and $\gamma$ is hyperparameter in focal loss. Note that, the focal loss is a widely-used loss for class imbalance  issue~\cite{zhang2019whole,zhang2021deep}.
To further improve the classification performance, we encourage high-density compactness of intra-class samples and low-density separation of inter-class samples for all domains via entropy minimization~\cite{mangin2000entropy}:

\begin{equation}
\label{en-class}
\mathcal{L}_{en}(\theta_{f},\theta_{m},\theta_{y}) = -\frac{1}{n_t + n_s} \sum_{i=1}^{n_t + n_s}\mathbf{\hat{y}}_{i}^{\top}\mathrm{log}(\mathbf{\hat{y}}_{i}).
\end{equation}
Based on the above, we summarize the overall classification loss $\mathcal{L}_{c}$ as follows:
\begin{equation}
    \mathcal{L}_{c}(\theta_{f},\theta_{m},\theta_{y}) = \mathcal{L}_{fo}(\theta_{f},\theta_{m},\theta_{y}) + \eta \mathcal{L}_{en}(\theta_{f},\theta_{m},\theta_{y}),
\end{equation}
where $\eta$ is a hyperparameter to trade-off between the two losses.

\section{Experiments}


\newl
\noindent{\bf{Dataset.}}
In this paper, we conduct our experiments on one anterior segment optical coherence tomography (AS-OCT) dataset, provided by	Zhongshan Ophthalmic Center. Such a dataset consists of a well-labeled source domain (\textbf{AS-I}) and two unlabeled target domains (\textbf{AS-II} and \textbf{AS-III}).
The data from different domains are acquired from different OCT devices and/or different countries. 
The statistics of the dataset are shown in Table~\ref{tab:data}.

\newl
\noindent{\bf{Compared methods.}}
We compare our proposed \method with one baseline (\textbf{Source-only}), several state-of-the-art unsupervised domain adaptation methods (\textbf{DSN}~\cite{bousmalis2016domain}, \textbf{DANN}~\cite{ganin2014unsupervised}, \textbf{DMAN}~\cite{zhang2019whole} and \textbf{ToAlign}~\cite{Wei2021ToAlignTA}), and one advanced multi-target domain adaptation methods for natural images (\textbf{MTDA}~\cite{gholami2020unsupervised}).
The baseline Source-only is trained only on the labeled source  domain.

\newl
\noindent{\bf{Implementation details.}} 
We implement our proposed method based on PyTorch~\cite{paszke2019pytorch}.
For a fair comparison, we use res2net~\cite{gao2019res2net} as the feature extractor in all considered methods.
(one can also use other DNN models, \eg ResNet~\cite{he2016deep} and MobileNetV2~\cite{sandler2018mobilenetv2}).
For all compared methods, we keep the same hyperparameters as the original paper. Note that we conduct pair-wise domain adaptation for each target domain separately when implementing unsupervised domain adaptation methods.
For \method, both classifier and domain discriminator consist of three fully connected layers.
In the training process, we use an SGD optimizer with a learning rate of 0.001 to train the network.
For the trade-off parameters, we set $\lambda=1.0$, $\eta=0.1$  and $\alpha=0.03$ through cross-validation.
Following~\cite{lin2017focal}, we set $\gamma=2.0$ for the focal loss.  
Following~\cite{fu2017segmentation}, we cut the images in half as the input of the network.

\begin{table}[tp]
  \caption{ 
  \cyf{Statistics of the AS-OCT dataset.}}
  \centering 
  \begin{threeparttable}
  \renewcommand\arraystretch{1.0}
  \renewcommand{\tabcolsep}{2.0pt}
  \scalebox{0.92}{
  \label{tab:data}
    \begin{tabular}{cccccccccc}
    \toprule
    \multirow{2}{*}{Domain} & \multirow{2}{*}{Data} &\multirow{2}{*}{Country}&\multirow{2}{*}{Device}&
    \multicolumn{3}{c}{Training set}&\multicolumn{3}{c}{ Test set}\cr
    \cmidrule(lr){5-7} \cmidrule(lr){8-10}
    & & & & \#Narrow & \#Open & \#Total & \#Narrow & \#Open & \#Total\cr
    \midrule
    Source & AS-I & China & CASIA-I & 3,006 & 6,024 & 9,030 & 790 & 1412 & 2,202\cr
    Target I & AS-II & China & CASIA-II& 62 & 464 & 526 & 64 & 464 & 526\cr
    Target II & AS-III&Singapore& CASIA-I& 416 & 1,406 & 1,822 & 418 & 1,406 & 1,824\cr
    \bottomrule
    \end{tabular}
    }
    \end{threeparttable}
\end{table}


\subsection{Comparisons with State-of-the-art Methods}
We compare our \method with several state-of-the-art methods in terms of two metrics, \ie accuracy and AUC~\cite{lobo2008auc}.
From Table~\ref{tab:result}, all domain adaptation methods perform better than Source-only, which verifies the contribution of unsupervised domain adaptation.
Moreover, our proposed \method outperforms all pair-wise unsupervised domain adaptation methods (\ie DSN, DANN, DMAN and DANN+ToAlign).
The result indicates that those pair-wise domain adaptation methods may fail to alleviate the discrepancies among multiple domains since they focus on pair-wise adaptation and cannot capture correlation among domains effectively.
In addition, \method also outperforms MTDA in terms of both metrics, which demonstrates the superiority of our proposed method in handling multi-target domain adaptation for angle closure classification. In \method, the multi-scale scheme helps to extract multi-scale domain-invariant features of AS-OCT images for angle closure  classification.  
Note that the performance of MTDA is worse than pair-wise domain adaptation methods. Such poor performance of MTDA results from the poor reconstruction which includes a lot of noise for fine-grained anterior chamber angle in AS-OCT images.


\begin{table}[t]
  \centering
  \caption{
 Comparisons of six methods in accuracy and AUC on two target domains.
  }
  \label{tab:result}
  \renewcommand\arraystretch{1.0}
  \renewcommand{\tabcolsep}{8.0pt}
  \scalebox{0.88}{
    \begin{tabular}{ccccccccc}
    \toprule
    \multirow{2}{*}{Method}&
    \multicolumn{2}{c}{AS-II}&\multicolumn{2}{c}{AS-III }&\multirow{2}{*}{Mean Acc.}&\multirow{2}{*}{Mean AUC}\cr
    \cmidrule(lr){2-3} \cmidrule(lr){4-5}
    &Acc.&AUC&Acc.&AUC\cr
    \midrule
    Source-only&0.638&0.834&0.737&0.911&0.688&0.872\cr
    DANN~\cite{ganin2014unsupervised} &0.723&0.866&0.836&0.912&0.780&0.889\cr
    DSN~\cite{bousmalis2016domain} &0.762&0.888&0.848&0.915&0.805&0.901\cr
    DMAN~\cite{zhang2019whole} &0.786&0.906&0.834&0.922&0.810&0.914\cr
    DANN+ToAlign~\cite{Wei2021ToAlignTA} &0.842&0.685&0.688&0.607&0.765&0.646\cr
    MTDA~\cite{gholami2020unsupervised} &0.667&0.791&0.735&0.806&0.701&0.799\cr
    \midrule
    \method (ours)&{\bf 0.856}&{\bf 0.914}&{\bf0.869}&{\bf 0.928}&{\bf 0.862}&{\bf 0.921}\cr
    \bottomrule
    \end{tabular}
    }
\end{table}

\begin{table}[t]
  \centering
  \caption{
  Effect of the filter size in the multi-scale module. The variant \method-S$k$ represents the convolutional layers in all three branches of the multi-scale module use the same filter size $k$, where $k$=1, 3, 5.}
  \label{tab:ablation}
  \renewcommand\arraystretch{1.0}
  \renewcommand{\tabcolsep}{8.0pt}
  \scalebox{0.9}{
    \begin{tabular}{ccccccccc}
    \toprule
    \multirow{2}{*}{Method}&
    \multicolumn{2}{c}{ AS II}&\multicolumn{2}{c}{ AS III}&\multirow{2}{*}{Mean Acc.}&\multirow{2}{*}{Mean AUC}\cr
    \cmidrule(lr){2-3} \cmidrule(lr){4-5}
    &Acc.&AUC&Acc.&AUC\cr
    \midrule
    \method-S1 &0.737 &0.847&0.807 &0.909  & 0.772&0.878 \cr
    \method-S3 &0.756 &0.913&0.748 &0.916  &0.752 &0.914 \cr
    \method-S5 &0.757 &0.902&0.792 &0.921 &0.775 &0.914 \cr
    \method&{\bf 0.856}&{\bf 0.914}&{\bf0.869}&{\bf 0.928}&{\bf 0.862}&{\bf 0.921}\cr
    \bottomrule
    \end{tabular}
    }
\end{table}

\begin{table}[t]
\caption{\label{tab:ablation2}Ablation study for the losses (\ie $\mathcal{L}_{fo}$, $\mathcal{L}_{d}$ and $\mathcal{L}_{en}$). Note that $\mathcal{L}_{ce}$ denotes the cross-entropy loss for angle closure classification.}
\vspace{-0.2in}
    \begin{center}
    \renewcommand{\tabcolsep}{4.0pt}
    \scalebox{0.95}{
         \begin{tabular}{ccccc|cccccc}
         \toprule
         \multirow{2}{*}{Backbone}&\multirow{2}{*}{$\mathcal{L}_{ce}$} & \multirow{2}{*}{$\mathcal{L}_{fo}$}& \multirow{2}{*}{$\mathcal{L}_{d}$} & \multirow{2}{*}{$\mathcal{L}_{en}$} & \multicolumn{2}{c}{ AS II}&\multicolumn{2}{c}{ AS III}&\multirow{2}{*}{Mean Acc.}&\multirow{2}{*}{Mean AUC} \\
         \cmidrule(lr){6-7} \cmidrule(lr){8-9}
          && & & & Acc. & AUC & Acc. & AUC \cr
         \midrule
         \cmark & \cmark & & &  &0.784 &0.826 &0.780&0.903&0.782&  0.865 \\
         \cmark && \cmark &  &  &0.828 &0.843 &0.814&0.894&0.821&  0.868 \\
         \cmark& & \cmark & \cmark &  &0.856&0.876&0.841&0.916&0.848& 0.896 \\
         \cmark& & \cmark & \cmark & \cmark &\textbf{0.856} &\textbf{0.914}&\textbf{0.869}&\textbf{0.928}&\textbf{0.862}&  \textbf{0.921}\\
         \bottomrule
         \end{tabular}
         }
    \end{center}
\end{table}

\begin{table}[t]
  \centering
  \caption{
 Influence of the trade-off parameter $\alpha$ on AUC performance of our method. The value of $\alpha$ is selected among [0.0003, 0.003, 0.03, 0.3],  while fixing other parameters.
  }
  \label{tab:result2}
  \renewcommand\arraystretch{0.9}
  \renewcommand{\tabcolsep}{25.0pt}
    \begin{tabular}{cccc}
    \hline 
    $\alpha$& AS-II &AS-III&Mean AUC\\
    \hline
    $3e{-04}$&0.828&0.910&0.869\cr
    $3e{-03}$ &0.909&0.892&0.901\cr
    $3e{-02}$&{\bf 0.914}&{\bf 0.928}&{\bf 0.921}\cr
    $3e{-01}$ &0.813&0.843&0.828\cr
    \hline
\end{tabular}
\end{table}

\begin{table}[!h]
  \centering
  \caption{
 Influence of the trade-off parameter $\eta$ on AUC performance of our method. The value of $\eta$ is selected among [0.001, 0.01, 0.1, 1], while fixing other parameters.}
  \label{tab:result3}
  \renewcommand\arraystretch{0.9}
  \renewcommand{\tabcolsep}{25.0pt}
    \begin{tabular}{cccc}
    \hline 
    $\eta$ &AS-II &AS-III&Mean AUC\\
    \hline
    $1e{-03}$&0.791& 0.871& 0.831\cr
    $1e{-02}$ &0.825&0.876&0.851\cr
    $1e{-01}$ &{\bf 0.914}& {\bf 0.928}&{\bf 0.921}\cr
    $1$ &0.884&0.922&0.903\cr
    \hline
\end{tabular}
\end{table}

\subsection{Ablation Studies} 
\subsubsection{The Effectiveness of Multi-scale Module.}
To verify the effectiveness of the proposed multi-scale module, we compare our method with three variants, namely \textbf{\method-S1}, \textbf{\method-S3} and \textbf{\method-S5}.
The variant \textbf{\method-S$k$} uses the same filter size $k$ in all three branches of convolutional layers in the multi-scale module.
In this case, the feature maps in each variant are extracted at the same scale.
From Table~\ref{tab:ablation}, \method achieves better performance than all three variants in terms of two target domains.
The results demonstrate the superiority of the proposed multi-scale module. 

\subsubsection{Ablation Studies on Difference Losses.} To investigate the effect of all losses ($\mathcal{L}_{fo}$, $\mathcal{L}_{d}$ and $\mathcal{L}_{en}$), we evaluate the model optimized by different losses. From Table~\ref{tab:ablation2}, our method with $\mathcal{L}_{fo}$ performs better than that with $\mathcal{L}_{ce}$, verifying that the focal loss helps to handle the class-imbalanced issue. When introducing $\mathcal{L}_{d}$, the performance improves a lot, which indicates that such a loss succeeds in alleviating domain discrepancies among domains. By combining all the losses, we obtain the best result. Such a result demonstrates that encouraging high-density compactness of intra-class
samples and low-density separation of inter-class samples further facilitates the classification of angle closure.

\begin{table}[t]
  \centering
  \caption{
 Influence of the trade-off parameter $\lambda$ on AUC performance of our method. The value of $\lambda$ is selected among [0.001, 0.01, 0.1, 1], while fixing other parameters.
  }
  \label{tab:result4}
  \renewcommand\arraystretch{0.9}
  \renewcommand{\tabcolsep}{25.0pt}
    \begin{tabular}{cccc}
    \hline 
    $\lambda$&AS-II &AS-III &Mean AUC\\
    \hline
    $1e{-03}$ & 0.663 & 0.671 & 0.667\cr
    $1e{-02}$ &0.709 & 0.647 &0.678 \cr
    $1e{-01}$ &0.659 &0.815 &0.737 \cr
    $1$ &{\bf 0.914}& {\bf 0.928} &{\bf 0.921}\cr
    \hline
\end{tabular}
\end{table}

\subsection{Influence of Hyper-parameter}
In this section, we investigate the impact of the hyper-parameters $\alpha$, $\eta$ and $\lambda$. We evaluate one parameter a time while fixing other parameters. As shown in Tables \ref{tab:result2}, \ref{tab:result3} and \ref{tab:result4}, our method achieves the best performance when setting $\alpha = 0.03$, $\eta = 0.1$ and $\lambda = 1$. To some extent, our method is non-sensitive to hyper-parameters. Moreover, it is crucial to set a reasonable classification loss weight which helps to classify angle closure effectively.

\section{Conclusion}
In this paper, we have proposed a novel Multi-scale Multi-target Domain Adversarial Network (\method) for angle closure classification.
\method aims to transfer a deep model learned on one labeled source domain to multiple unlabeled target domains. 
To be specific, we devise a multi-scale module to extract features regarding both local and global information.
By performing multi-domain adversarial learning at different scales, \method is able to extract domain-invariant features and effectively classify angle closure in multiple target domains.
Extensive experiments demonstrate the effectiveness of our proposed method.

\section*{Acknowledgments}
This work was partially supported by Key Realm R\&D Program of Guangzhou (202007030007), National Natural Science Foundation of China (NSFC) 62072190 and Program for Guangdong Introducing Innovative and Enterpreneurial
Teams 2017ZT07X183.

%
%
%
\bibliographystyle{splncs04}
\bibliography{mybi}

\end{document}